\documentclass[nohyperref]{article}

\usepackage{microtype}
\usepackage{graphicx}
\usepackage{subfigure}
\usepackage{booktabs} % for professional tables

\usepackage{hyperref}

\usepackage[accepted]{icml2022}

\usepackage{amsmath}
\usepackage{amssymb}
\usepackage{mathtools}
\usepackage{amsthm}

\theoremstyle{plain}
\newtheorem{thm}{Theorem}[section]

\newtheorem{prop}[thm]{Proposition}
\newtheorem{rem}{Remark}[section]

\theoremstyle{definition}
\newtheorem{defn}{Definition}[section]

\newtheorem{exmp}{Example}[section]

\usepackage[capitalize,noabbrev]{cleveref}

\theoremstyle{plain}

\theoremstyle{definition}

\theoremstyle{remark}

\usepackage[textsize=tiny]{todonotes}

\icmltitlerunning{A Geometric Perspective on Autoencoders}

\begin{document}

\twocolumn[
\icmltitle{A Geometric Perspective on Autoencoders}




\begin{icmlauthorlist}
\icmlauthor{Yonghyeon Lee}{yyy}
% \icmlauthor{Firstname2 Lastname2}{equal,yyy,comp}
% \icmlauthor{Firstname3 Lastname3}{comp}
\end{icmlauthorlist}

\icmlaffiliation{yyy}{Korea Institute for Advanced Study, Center for AI and Natural Sciences, Seoul, South Korea}

\icmlcorrespondingauthor{Yonghyeon Lee}{ylee@kias.re.kr}

\icmlkeywords{Machine Learning, ICML}

\vskip 0.3in
]



\printAffiliationsAndNotice{}  % leave blank if no need to mention equal contribution

\begin{abstract}
This paper presents the geometric aspect of the autoencoder framework, which, despite its importance, has been relatively less recognized. 
Given a set of high-dimensional data points that approximately lie on some lower-dimensional manifold, an autoencoder learns the \textit{manifold} and its \textit{coordinate chart}, simultaneously. 
This geometric perspective naturally raises inquiries like ``Does a finite set of data points correspond to a single manifold?'' or ``Is there only one coordinate chart that can represent the manifold?''. The responses to these questions are negative, implying that there are multiple solution autoencoders given a dataset. Consequently, they sometimes produce incorrect manifolds with severely distorted latent space representations. In this paper, we introduce recent geometric approaches that address these issues.
\end{abstract}

\section{Introduction}
\label{intro}
Observations from real-world problems are often high-dimensional, 
i.e., a large number of variables is needed to numerically represent the observed data.
{\it \bf Manifold hypothesis} assumes that high-dimensional data 
lie approximately on some lower-dimensional manifold, suggesting that a 
data set that appears to initially require many variables to describe can 
actually be described by a comparatively small number of variables; 
below shows one intuitive example.
\begin{exmp}{\bf Rotating face image manifold.} 
There is a sequence of pictures of a person turning his head from left to right 
as shown in Figure~\ref{fig:facemanifold}, where each image size is 
$3\times 728 \times 1280$. These images can be initially viewed as elements 
of the high-dimensional image data space $\mathbb{R}^{3 \times 728 \times 1280}$, 
however, they clearly do not fill up the entire image space, but rather form 
a lower-dimensional subspace. We need only one variable 
to represent these images, e.g., the angle of the person's head, meaning that 
the set of images can be interpreted as lying on a one-dimensional subspace 
or a curve as shown in Figure~\ref{fig:facemanifold}. 
\begin{figure}[!t]
  \centering
  \includegraphics[width=\linewidth] {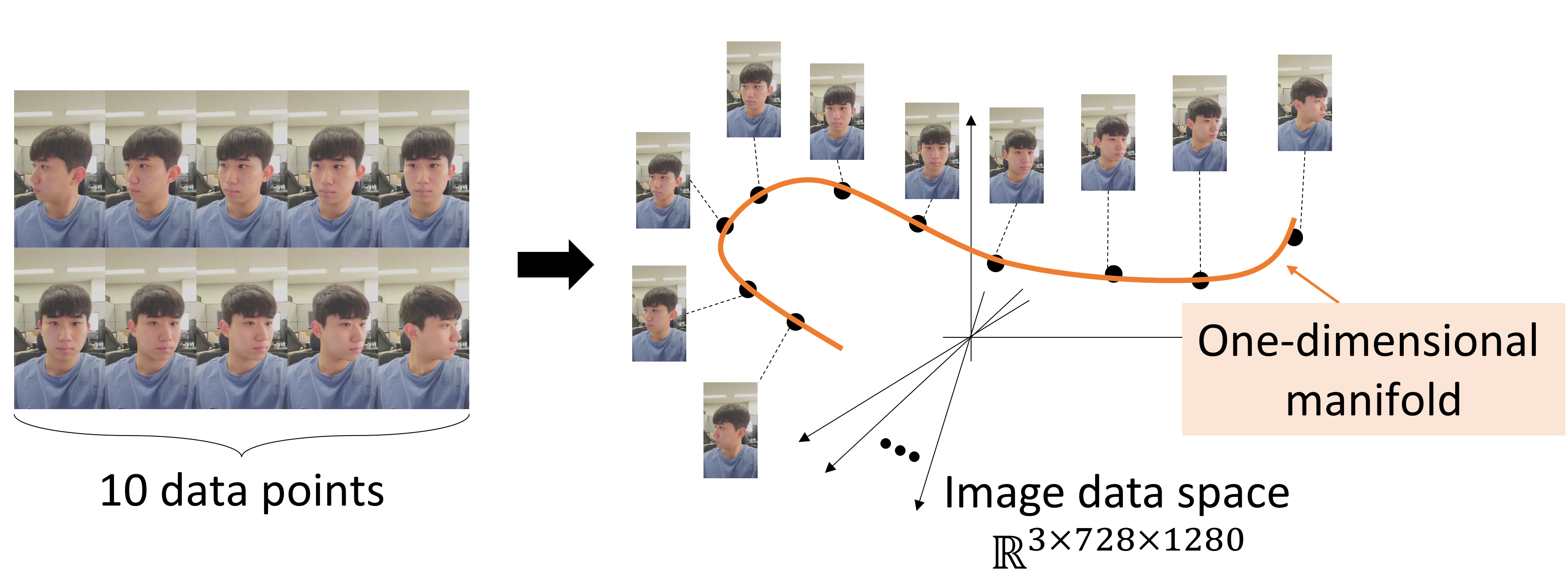}
  \vspace{-10pt}
  \caption{A set of face images, where a person turns his head from left to right, 
  approximately lie on a one-dimensional manifold or a curve.}
  \vspace{-10pt}
  \label{fig:facemanifold}
\end{figure}
\end{exmp}

\begin{figure}[!t]
  \centering
  \includegraphics[width=0.9\linewidth] {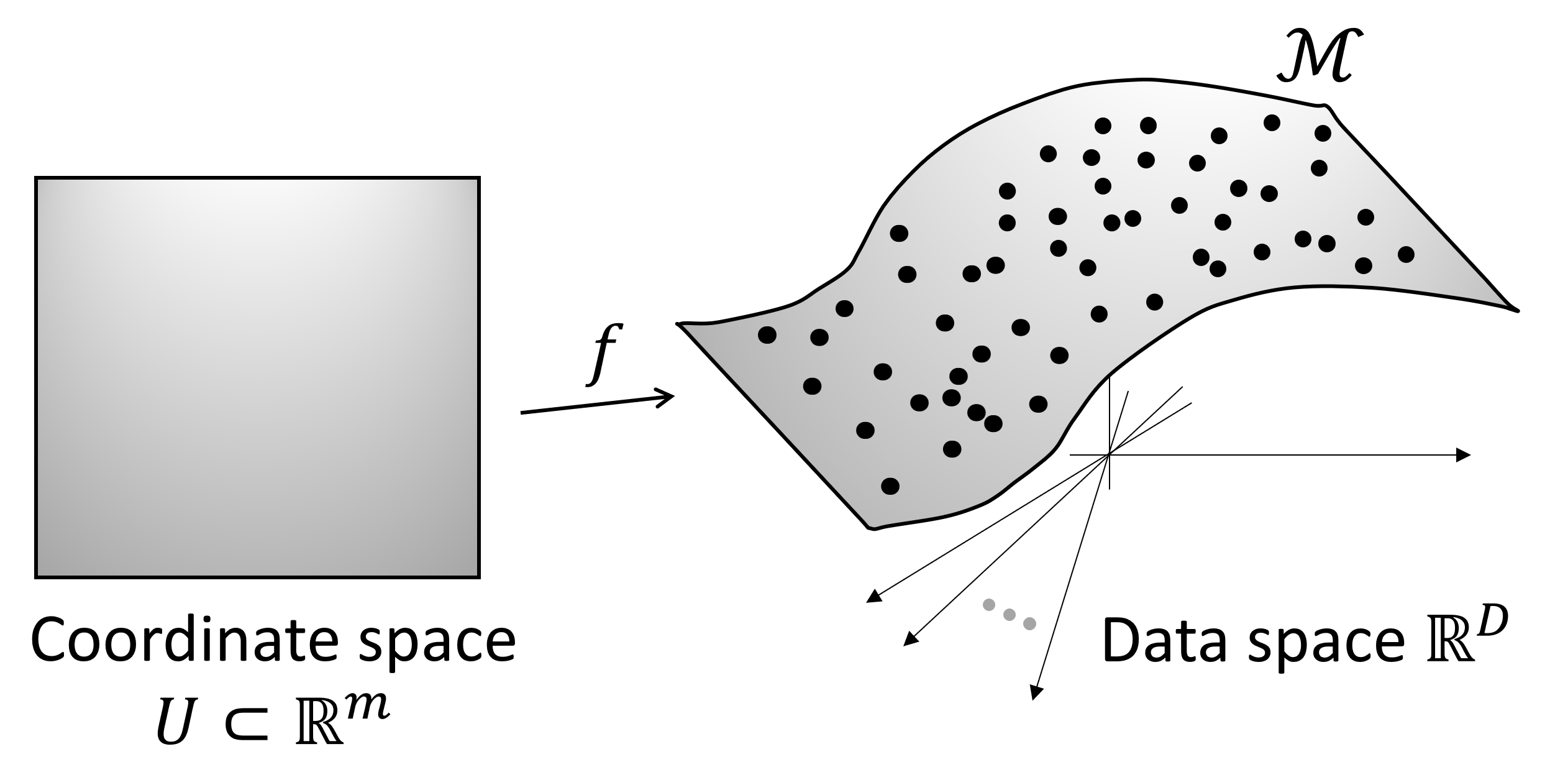}
  \vspace{-10pt}
  \caption{An illustration of the manifold ${\cal M}$ and coordinate chart $f:U\to{\cal M}$.}
  \vspace{-10pt}
  \label{fig:mlandchart}
\end{figure}

Under the manifold hypothesis, suppose a finite set of data points $\{x_i \in \mathbb{R}^{D}\}$
approximately lie on some manifold of dimension $m<D$, denoted by ${\cal M}$.
In other words, the support of the underlying data distribution $p_{X}(x)$ is contained in ${\cal M}$, i.e., $\text{supp}(p_{X}) \subset {\cal M}$. 
Then, the \textbf{manifold representation learning} problem consists of the following two components:
(i) identifying the manifold ${\cal M} \subset \mathbb{R}^{D}$ and (ii) finding a coordinate chart, i.e., a continuous and invertible map $f:U \to {\cal M}$ for some subset $U \subset \mathbb{R}^{m}$ (see Figure~\ref{fig:mlandchart}). 

The \textbf{autoencoder} framework provides an effective way of learning the manifold and coordinate chart simultaneously~\cite{kramer1991nonlinear}, together with the deep learning techniques used for approximating arbitrary complex functions~\cite{lecun2015deep}. 
The core idea is to learn two neural networks, an encoder $g_{\phi}:\mathbb{R}^{D}\to\mathbb{R}^{m}$ and a decoder $f_{\theta}:\mathbb{R}^{m}\to\mathbb{R}^{D}$, in a way that the composition of them reconstructs all the given data points $x_i\in \mathbb{R}^{D}$, i.e., $f \circ g(x_i) \approx x_i$, by solving the following reconstruction loss minimization problem:
\begin{equation}
\min_{\theta,\phi} \sum_{i=1}^{N}\|x_i-f_{\theta}(g_{\phi}(x_i))\|^2.
\end{equation}

\begin{figure}[!t]
    \centering
    \includegraphics[width=\linewidth]{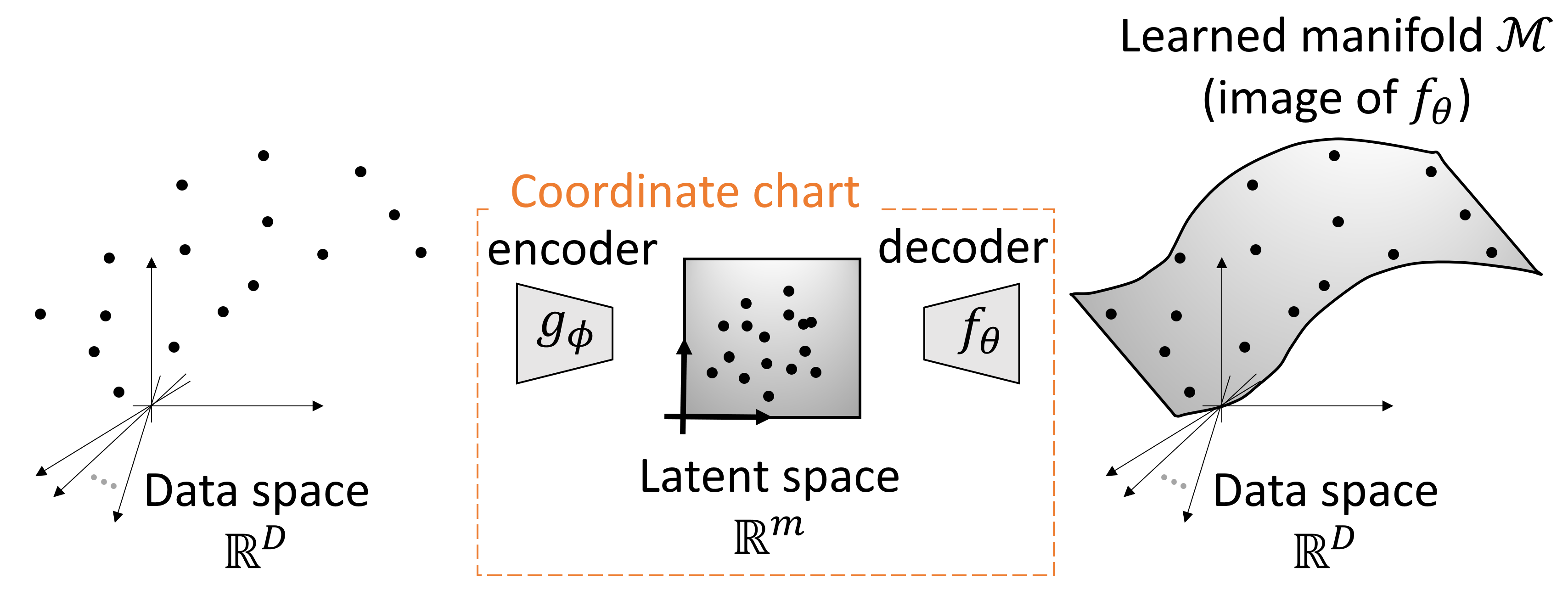}
    \vspace{-10pt}
    \caption{An autoencoder-based simultaneous manifold and coordinate chart learning}
    \label{fig:aemanifold}
    \vspace{-10pt}
\end{figure}

Provided that the loss approaches near zero, the data points become to lie on the image of the decoder $f_{\theta}$, 
i.e., $x_i \in \text{image}(f_{\theta})$. 
Under some suitable conditions, the image of the decoder is a differentiable manifold, and it can be seen that $f_{\theta}$ has learned the manifold ${\cal M}$:
\begin{rem}
    If $f_{\theta}:\mathbb{R}^{m} \to \mathbb{R}^{D}$ is a smooth injection and its Jacobian is full rank, then the image of $f_{\theta}$ is a differentiable manifold embedded in $\mathbb{R}^{D}$. 
\end{rem}
The encoder $g_{\phi}$ maps the data points $x_i$ to some low-dimensional latent vectors in $\mathbb{R}^{m}$, and 
the subset $U$ can be found by fitting a probability density function to $\{g_{\phi}(x_i)\}$ and taking its support.

For ease of discussion, throughout, we will assume that the manifold ${\cal M}$ is homeomorphic to $\mathbb{R}^{m}$, i.e., ${\cal M}$ is continuously deformable to $\mathbb{R}^{m}$, and consider the entire image of $f_{\theta}:\mathbb{R}^{m} \to \mathbb{R}^{D}$ as the \textbf{learned manifold} (or \textbf{decoded manifold}); see Figure~\ref{fig:aemanifold}. We denote by ${\cal M}$  this decoded manifold by a slight abuse of notation, and the manifold is said to be explicitly parametrized by $f_{\theta}$.
Then, informally, we can view $g_{\phi}:{\cal M} \to \mathbb{R}^{m}$ as an approximate inverse of $f_{\theta}:\mathbb{R}^{m} \to {\cal M}$. The encoder and decoder with the latent space $\mathbb{R}^{m}$ together take the role of the \textbf{coordinate chart}. 

In vanilla autoencoders, there are two fundamental issues that we aim to address: (i) they often learn incorrect manifolds that overfit to training data or have the wrong local connectivity and geometry, and (ii) they learn geometrically distorted latent representations where geometric quantities such as the lengths, angles, and volumes in the data manifold are not preserved. Below, we will take a closer look at the following issues one by one.

\subsection{Wrong Manifold}
\begin{figure}[!t]
  \centering
  \includegraphics[width=\linewidth] {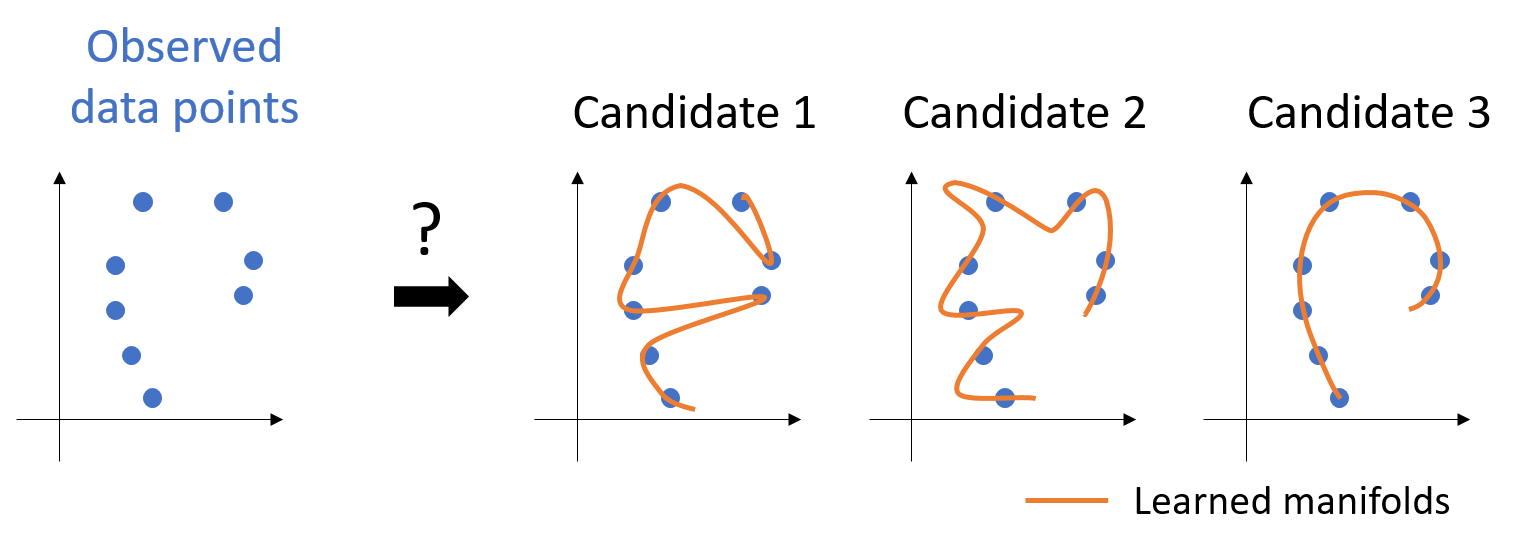}
  \vspace{-10pt}
  \caption{An illustration of the fundamental wrong manifold issue in autoencoder-based manifold learning. Given blue data points, there are many manifolds where the given data points perfectly lie.}
  \vspace{-10pt}
  \label{fig:wmissue}
\end{figure}

In the traditional autoencoder training setting, we are given a set of finite data points $\{x_i\in\mathbb{R}^{D}\}_{i=1}^{N}$. 
Assuming the data points are clean and perfectly lie on the ground-truth data manifold $\mathcal{M}$, the primary condition for a correct coordinate system $f:\mathbb{R}^{m} \to \mathbb{R}^{D}$ to satisfy is the following:  
\begin{equation}
x_i \in \mathcal{M} \subset \mathbb{R}^{D} \text{  for all  } i \Longleftrightarrow x_i \in \text{image}(f) \text{  for all  } i.
\end{equation}
However, finding such a map $f$ is fundamentally ill-posed, meaning that there are infinitely many mappings $f$ that satisfy the above condition.
For example, as shown in Figure~\ref{fig:wmissue}, assume that there exists a ground-truth one-dimensional manifold embedded in the two-dimensional data space and we are given a set of finite two-dimensional data points (blue points). 
There are many manifolds (orange manifolds) where the given data points perfectly lie; thus, we cannot specify the solution manifold without further assumptions on the manifold or decoder $f$.

\subsection{Distorted Latent Space}
\begin{figure}[!t]
  \centering
  \includegraphics[width=1\linewidth] {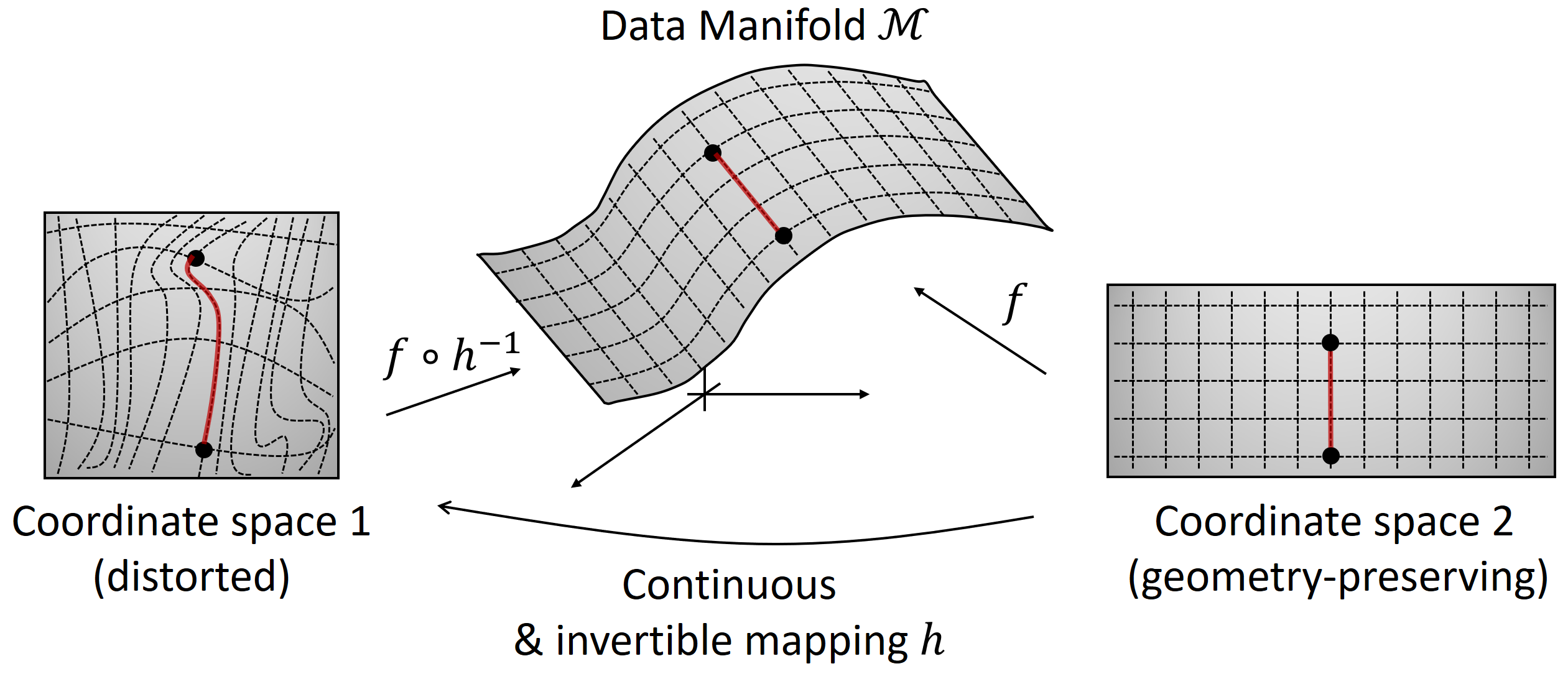}
  \vspace{-10pt}
  \caption{There are many coordinate systems for a data manifold $\mathcal{M}$; some produce geometrically distorted coordinate spaces.}
  \vspace{-10pt}
  \label{fig:distortedissue}
\end{figure}

Given a ground-truth data manifold $\mathcal{M}$, the problem of finding a coordinate system for $\mathcal{M}$ is again fundamentally ill-posed because there exist infinitely many coordinate systems. 
For example, if $f:\mathbb{R}^{m}\to\mathcal{M}$ is one coordinate system, then for any continuous and invertible map $h:\mathbb{R}^{m} \to \mathbb{R}^{m}$ the composition map $f \circ h^{-1}:\mathbb{R}^{m} \to \mathcal{M}$ is also another coordinate system (Figure~\ref{fig:distortedissue}). 
Recall the reconstruction error objective function in the vanilla autoencoder $\sum_i \|x_i-f \circ g (x_i)\|^2$, if $f^*$ and $g^*$ are solutions, then for any continuous and invertible map $h:\mathbb{R}^{m} \to \mathbb{R}^{m}$ the composition maps $f^* \circ h^{-1}$ and $h \circ g^*$ are also solutions.
Therefore, we cannot favor one coordinate system over the others without further assumptions on the mappings.

\begin{figure}[!t]
  \centering
  \includegraphics[width=0.8\linewidth] {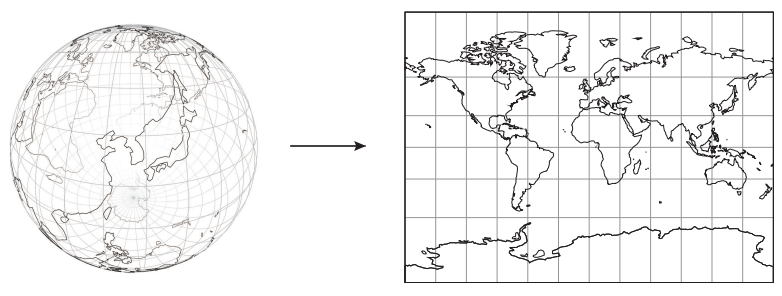}
  \caption[Map-making: finding a mapping from a sphere to a Cartesian plane.]{Map-making: finding a mapping from a sphere to a Cartesian plane, adopted from Figure 1.5 in~\cite{2019riemannian}.}
  \label{fig:map_making}
\end{figure}

\begin{figure}[!t]
  \vspace{10pt}
  \centering
  \includegraphics[width=0.8\linewidth] {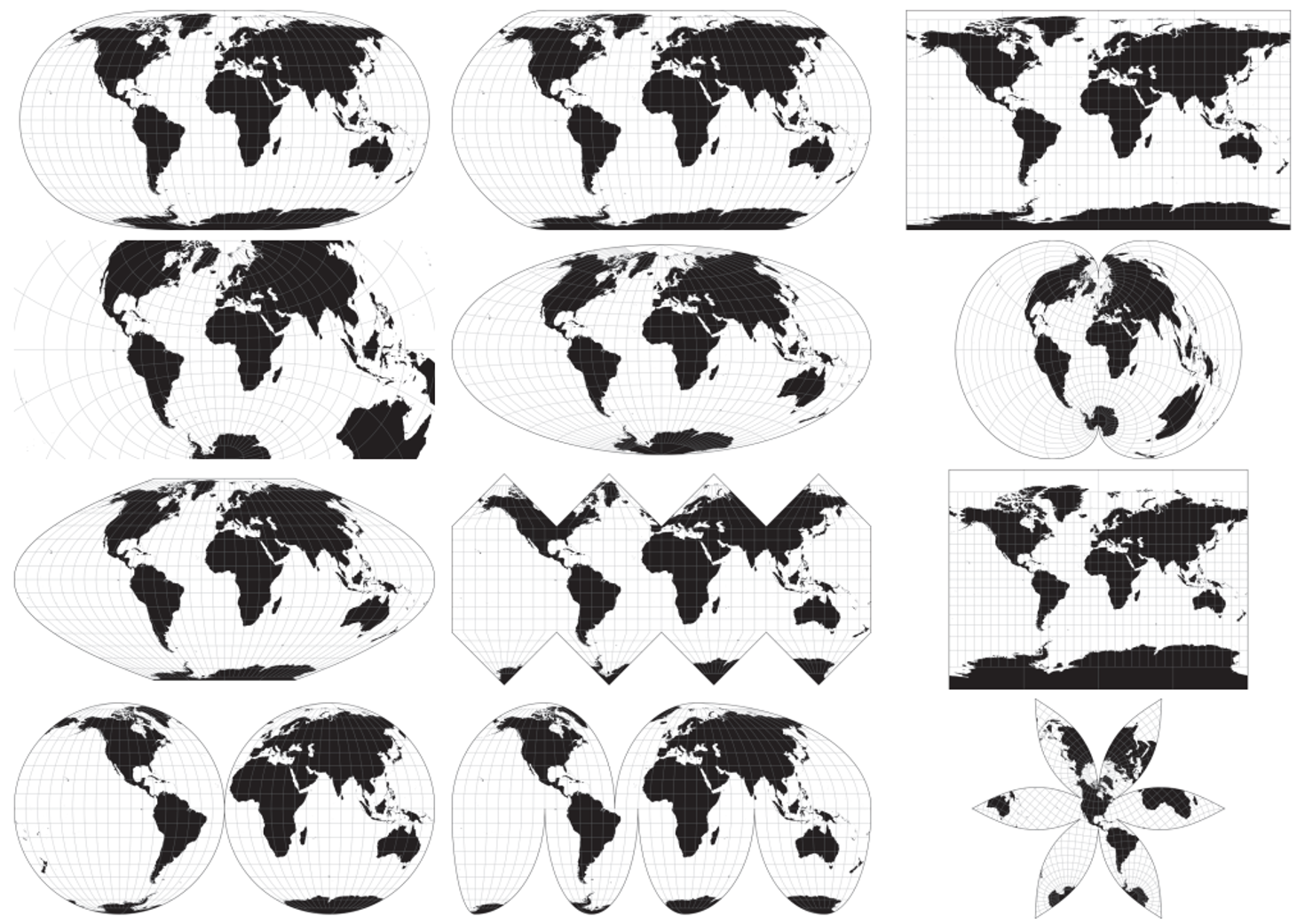}
  \caption[Maps of the earth.]{Maps of the earth, adopted from Figure 1.6 in~\cite{2019riemannian}.}
  \label{fig:maps_earth}
\end{figure}

\begin{figure}[!t]
  \vspace{10pt}
  \centering
  \includegraphics[width=0.8\linewidth] {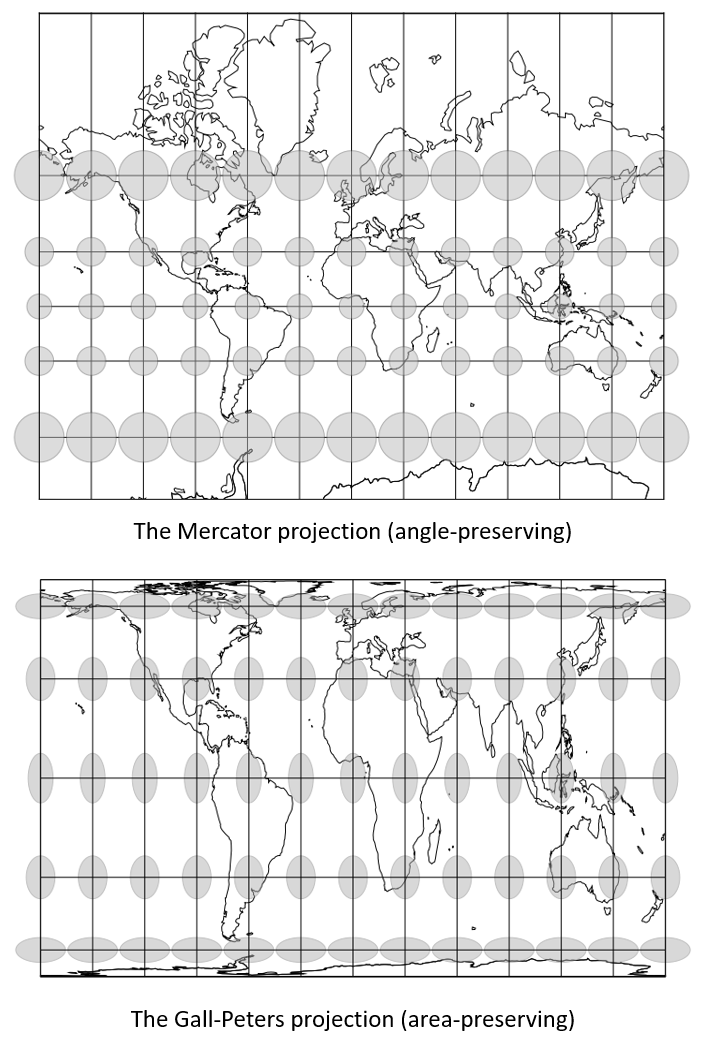}
  \caption[Two examples of maps of the earth.]{Two examples of maps of the earth, adopted from Figure 1.7 in~\cite{2019riemannian}.}
  \label{fig:mer_gall}
\end{figure}

This is closely related to the classical map-making problem, i.e., making a map of the earth, by finding a mapping from the surface of the earth approximated by a two-dimensional sphere $S^{2}$ to a two-dimensional Cartesian plane $\mathbb{R}^{2}$ (Figure~\ref{fig:map_making}).
As shown in Figure~\ref{fig:maps_earth}, a wide variety of maps of the earth exist, each of which is based on different projection methods.
In general, maps that are arbitrarily distorted are not preferred, rather it is designed to preserve specific intrinsic geometric quantities defined based on the purpose of the map.
For example, the Mercator projection preserves angles while the areas are distorted, and the Gall-Peters map preserves areas although the shapes are distorted (Figure~\ref{fig:mer_gall}). 

In this paper, we introduce recent geometric regularization methods that attempt to resolve these two issues. 
Before that, in section~\ref{sec:preliminary}, we introduce some geometric preliminaries required 
to understand the subsequent sections.
Section~\ref{sec:nrae}, which is based on~\cite{lee2021neighborhood}, introduces a method to resolve the 
wrong manifold issue by using additional information, a priori constructed neighborhood graph. 
We regularize the geometry and connectivity of the learned manifold to be close to those of the given neighborhood graph. 
Section~\ref{sec:mcae}, which is based on~\cite{lee2023explicit}, attempts to minimize the curvature of the learned manifold, 
prioritizing smooth manifolds over curved ones.
Section~\ref{sec:irae}, which is based on~\cite{lee2022regularized}, introduces a method to resolve the distorted latent space issue, by trying to regularize the decoder to be close to a geometry-preserving mapping.  
Besides ours, we note that there are many other autoencoder regularization methods developed from 
different perspectives~\cite{arvanitidis2017latent, shao2018riemannian, rifai2011contractive, kingma2013auto, tolstikhin2017wasserstein, makhzani2015adversarial, vincent2010stacked, chen2020learning, higgins2016beta, jang2022geometrically, chen2021neighborhood, duque2020extendable}.  

\section{Geometric Preliminaries}
\label{sec:preliminary}
Throughout, we denote the decoder by $f:\mathbb{R}^{m} \to \mathbb{R}^{D}$ such that 
$z \mapsto x=f(z)$, and consider its image as an 
$m$-dimensional differentiable manifold ${\cal M}$ embedded in $\mathbb{R}^{D}$.
The space $\mathbb{R}^{D}$ is called the ambient space. 
In this section, we discuss various geometric quantities of ${\cal M}$ and $f$.
We will denote the Jacobian of $f$ by $J_f \in \mathbb{R}^{D \times m}$. 

\subsection{Tangent Space and Riemannian Metric}

The decdoer Jacobian $J_{f}(z)$ spans the {\it tangent space} of the manifold ${\cal M}$ attached at $x=f(z)$ -- where we consider $x\in {\cal M}$ as an origin of the tangent vector space $T_{x} {\cal M}$ --, i.e, $T_{x} {\cal M} = \{x + J_f(z)v \ | \ v \in \mathbb{R}^{m}\} \subset \mathbb{R}^{D}$.
The $m$ column vectors in the Jacobian matrix $J_{f}(z)\in\mathbb{R}^{D\times m}$ can be viewed as a set of linearly independent basis vectors for the tangent space. 

We can give a geometric structure to a differentiable manifold ${\cal M}$ by assigning 
a {\it Riemannian metric}, i.e., inner products $\{\langle \cdot, \cdot \rangle_{x}\}$ for the tangent spaces $\{T_x {\cal M}\}$ that smoothly change with respect to $x$. 
One natural way to define the metric for ${\cal M}$ is to project the ambient space Riemannian metric to ${\cal M}$.
Let $H(x) \in \mathbb{R}^{D\times D}$ -- which is a positive-definite matrix -- be a Riemannian 
metric for the ambient space. 
Then, the projected metric to the embedded manifold ${\cal M}$ is defined as follows: 
for two tangent vectors $v,w \in T_{x} {\cal M} \subset \mathbb{R}^{D}$, $\langle v, w \rangle_x := v^T H(x) w$.

Given a decoder (or a coordinate chart), this projected metric can be expressed in local coordinates 
as $m\times m$ positive-definite matrices. Consider arbitrary two smooth curves $z_1(t)$ and $z_2(t)$ in $\mathbb{R}^{m}$ that intersect at $t=0$, i.e., $z=z_1(0)=z_2(0)$, and their corresponding curves in ${\cal M}$ denoted by $x_1(t)=f(z_1(t))$ and 
$x_2(t)=f(z_2(t))$. Differentiating both sides, we get $\dot{x}_1 = J_f(z) \dot{z}_1$ and $\dot{x}_2 = J_f(z) \dot{z}_2$. 
Note that $\dot{x}_1$ and $\dot{x}_2$ are two tangent vectors in $T_{x} {\cal M}$, and we can compute their inner product using the projected metric $\langle \dot{x}_1, \dot{x}_2 \rangle_x$; then, we get the following equality:
\begin{equation}
    \dot{x}_1^T H(x) \dot{x}_2 = \dot{z}_1^T J_f(z)^T H(f(z)) J_f(z) \dot{z}_2.
\end{equation}
The matrix $J_f(z)^T H(f(z)) J_f(z)$ is an $m\times m$ positive-definite matrix, 
which defines inner products between $\dot{z}_1$ and $\dot{z}_2$ for arbitrary curves $z_1(t),z_2(t)$ 
in the latent space.
It is called either a pull-back metric of $H$ by $f$, a 
coordinate representation of the projected metric of $H$ to ${\cal M}$, 
or sometimes a latent Riemannian metric.

Now that we have defined the geometry of ${\cal M}$ with the Riemannian metric, 
we can compute the length of a curve in ${\cal M}$. Let $x(t), t\in [0, 1]$ be a smooth curve in ${\cal M}$ 
and $z(t)$ be its coordinate representation, i.e., $x(t)=f(z(t))$. 
Given the projected metric $H$ to ${\cal M}$, the length of the curve can be computed 
by using the pull-back metric as follows:
\begin{equation}
\label{eq:len_of_curves}
    \text{Len}(x(t)):= \int_{0}^{1} \sqrt{\dot{z}^T J_f(z)^T H(f(z)) J_f(z) \dot{z}} \ dt.
\end{equation}
And, the energy of the curve is defined as 
\begin{equation}
    \text{Energy}(x(t)):= \int_{0}^{1} \dot{z}^T J_f(z)^T H(f(z)) J_f(z) \dot{z} \ dt.
\end{equation}
Given two points in ${\cal M}$, a {\it minimal geodesic} is defined as the energy-minimizing curve and {\it minimal geodesic distance} is defined as its length. 
Previous works~\cite{arvanitidis2017latent, shao2018riemannian, arvanitidis2020geometrically} computed geodesics and geodesic distances by solving geodesic equations or energy minimization problems.

\subsection{Manifold Curvature Measures}
This section explains the {\it extrinsic curvature measure} proposed in~\cite{lee2023explicit}.
Recall that the curvature of a curve is a measure of the local rate of change of the tangent vector: straight lines have zero curvature, while circles have constant curvatures. In the same way, the curvature of a manifold ${\cal M}$ in $\mathbb{R}^{D}$ can be defined as a measure of the local rate of change of the tangent space $T_x {\cal M}$. 

The tangent space $T_{x} {\cal M}$ corresponds to the range of the Jacobian, and we may choose $J_f(z)$ as a numerical representation of the tangent space. However, it is not coordinate invariant and thus not geometrically well-defined: given a coordinate transformation $z \mapsto \tilde{z}(z)$ and $f(z) =\tilde{f}(\tilde{z}(z))$, $J_f = \frac{\partial f}{\partial z} \mapsto J_{\tilde{f}} = J_f \frac{\partial z}{\partial \tilde{z}}$.  

Instead, we can represent the tangent space $T_{x} {\cal M}$ with a $D\times D$ orthogonal projection matrix in a coordinate-invariant way:
\begin{equation}
T_x {\cal M} = \text{range}(J_f (J_f^T J_f)^{-1} J_f^T),
\end{equation}
where $J_f(J_f^TJ_f)^{-1}J_f^T$ is the orthogonal projection matrix that projects a $D$-dimensional vector in $\mathbb{R}^{D}$ to $T_x {\cal M}$; this representation is invariant under the coordinate transformations:
\begin{prop}
    Given a coordinate transformation $z \mapsto \tilde{z}(z)$, $J_f (J_f^T J_f)^{-1} J_f^T = J_{\tilde{f}} (J_{\tilde{f}}^T J_{\tilde{f}})^{-1} J_{\tilde{f}}^T$.
\end{prop}

There is one-to-one correspondence between the set of $D\times D$ orthogonal projection matrices of rank $m$
and the set of $m$-dimensional linear subspaces in $\mathbb{R}^{D}$, which again validates 
our choice of the orthogonal projection matrix representation. More formally, this argument can be 
made by using the theory of the {\it Grassmann manifold}; see~\cite{lee2023explicit}. 

We denote the orthogonal projection matrix representation of $T_x {\cal M}$ by $\hat{T}=J_f(J_f^TJ_f)^{-1}J_f^T$. 
Given a curve $x(t)= f(z(t)) \in {\cal M}$, the local change of the tangent space along the curve can then be expressed in local coordinates as
\begin{equation}
\frac{d \hat{T}}{dt} = \sum_{i=1}^{m} \frac{\partial \hat{T}}{\partial z^i} \ \dot{z}^i \in \mathbb{R}^{D \times D}.
\end{equation}
And, its Frobenius norm is
\begin{equation}
	\|\frac{d \hat{T}}{dt}\|_F^2 = \sum_{i=1}^{m}\sum_{j=1}^{m} \text{Tr}(\frac{d\hat{T}}{dz^i}^T \frac{d\hat{T}}{dz^j}) \dot{z}^i \dot{z}^j;
\end{equation}
the resulting quadratic form is
\begin{equation}
	\|\dot{\hat{T}}\|_F^2 \ dt = \sum_{i=1}^{m} \sum_{j=1}^{m} c_{ij}(z) dz^i dz^j = dz^T C(z) dz,
\end{equation} 
where $dz = (dz^1, \ldots, dz^m)$, and the matrix $C(z) \in \mathbb{R}^{m \times m}$ is defined to be 
$c_{ij}(z)=\text{Tr}(\frac{d\hat{T}}{dz^i}^T \frac{d\hat{T}}{dz^j})$, which we call the {\it extrinsic curvature form}. 

Given the pull-back Riemannian metric $J^T_f H J_f$, we can define a local extrinsic curvature measure of ${\cal M} \subset \mathbb{R}^{D}$ at $z$ as follows:
\begin{equation}
\label{eq:excurvature}
\text{Tr}((J_f^T H J_f)^{-1} C) = \sum_{i,j}(J_f^T H J_f)^{-1}_{ij} \text{Tr}(\frac{d\hat{T}}{dz^i}\frac{d\hat{T}}{dz^j}),
\end{equation} 
which is a Dirichlet energy of the mapping between the Riemannian manifold ${\cal M}$ 
and the Grassmann manifold; see~\cite{lee2023explicit}.
This curvature measure is geometrically well-defined since 
it is invariant under the coordinate transformations:
\begin{prop}
    Given a coordinate transformation $z \mapsto \tilde{z}(z)$, $\text{Tr}((J_f^T H J_f)^{-1} C)$ is invariant.
\end{prop}

This is a local measure at $z\in \mathbb{R}^{m}$, and to define a global measure, 
we can either integrate the local curvature measure over some regions or compute 
the expectation over some probability measure in $\mathbb{R}^{m}$. 

This section exclusively deals with the extrinsic curvature measure that we have proposed. 
Although not introduced here, there are well-developed theories of intrinsic curvatures 
of Riemannian manifolds such as Riemann curvature, Ricci curvature, and scalar curvature; 
we refer to~\cite{do1992riemannian, fecko2006differential} for those who are interested. 
In~\cite{lee2023explicit}, algorithms for both extrinsic and intrinsic curvatures have been developed.

\subsection{A Hierarchy of Geometry-Preserving Mappings and Distortion Measures}
This section introduces a hierarchy of geometry-preserving mappings and 
coordinate-invariant distortion measures in~\cite{jang2020riemannian, lee2022regularized}.  
For simplicity, here we will restrict our attention to the decoder $f:\mathbb{R}^{m} \to \mathbb{R}^{D}$; 
formulations for mappings between general manifolds can be found in the original papers.
Throughout, we will assume that both latent and data spaces are Riemannian manifolds, 
where the latent space $\mathbb{R}^{m}$ is assigned the identity metric (i.e., Euclidean space), and 
the ambient data space is assigned the Riemannian metric $H(x)$. 

\subsubsection{A Hierarchy of Geometry-Preserving Mappings}
At the top of the hierarchy, an {\it isometry} is a mapping between two spaces that preserves distances and angles everywhere. For a linear mapping between two vector spaces equipped with inner products, 
an isometry preserves the inner product everywhere. In the case of a mapping between Riemannian manifolds, 
$f$ is an isometry if 
\begin{equation}
\label{eq:iso_condition}
    I = J_f(z)^T H(f(z)) J_f(z) \:\:\: \forall z \in \mathbb{R}^{m}.
\end{equation}
This equality condition is derived from the requirement that the $\text{Len}(x(t))$ in ${\cal M}$ expressed in the equation (\ref{eq:len_of_curves}) must be equal to the $\text{Len}(z(t))=\int_{0}^{1} \sqrt{\dot{z}^T \dot{z}} \ dt$ in $\mathbb{R}^{m}$ for all curve $x(t) \in {\cal M}$. 

Sometimes, requiring a map $f$ to be an isometry can be overly
restrictive; preserving only angles may be sufficient.  A {\it
conformal map} is a mapping that preserves angles but not
necessarily distances. Mathematically, $f$ is conformal (or angle-preserving) if
\begin{equation}
I = c(z)J_f(z)^T H(f(z)) J_f(z) \:\:\:
\forall z \in \mathbb{R}^{m},
\end{equation}
for some positive function $c:\mathbb{R}^{m} \to \mathbb{R}$.
The positive function is called the conformal factor.

A conformal map with a constant conformal factor, i.e., one in which
$c(z)$ is constant, sits one level below the isometric mapping and
is defined formally as any mapping $f$ for which a positive scalar
constant $c$ satisfying
\begin{equation}
\label{eq:scalediso_condition}
I = c J_f(z)^T H(f(z)) J_f(z) \:\:\:
\forall z \in \mathbb{R}^{m}
\end{equation}
can be found. Such a map not only preserves angles but also scaled
distances; for this reason, this mapping is referred to as the {\it scaled isometry}. 

\subsubsection{Distortion measures}
We begin by introducing another equivalent characterization of the isometry condition in (\ref{eq:iso_condition}). 
$f$ is an isometry if all the eigenvalues of $J_f^T(z) H(f(z)) J_f(z)$ are equal to 1 for all $z \in \mathbb{R}^{m}$. 
Then, denoting the eigenvalues by $\{\lambda_i(z)\}$, we can define a {\it local distortion measure} of $f$, which 
measures how far the mapping $f$ from being a local isometry, as follows:
\begin{equation}
    \sum_{i=1}^{m} (1-\lambda_i(z))^2.
\end{equation}
This local measure can be integrated over some probability measure $\nu$ in $\mathbb{R}^{m}$ 
to define a {\it global distortion measure}:
\begin{equation}
    \int_{\mathbb{R}^{m}} \sum_{i=1}^{m} (1- \lambda_i(z))^2 \ d \nu(z).
\end{equation} 
Since the influence of the measure is limited to the support of $\nu$, 
if the global distortion measure of $f$ is zero, 
then $f$ is an isometry with respect to the support of $\nu$. 
For simplicity, we omit the expression `with respect to the support of $\nu$' whenever it is 
clear from the context.

One may naively want to consider $\|I-J_f^THJ_f\|_F^2$ as a local distortion measure, 
but such a measure is not coordinate-invariant.
Our distortion measure defined with the eigenvalues is {\it coordinate-invariant} and thus 
is geometrically well-defined. We refer to~\cite{lee2022regularized, jang2020riemannian} for  
a general strategy of constructing coordinate-invariant functionals on Riemannian manifolds
and general formulations of coordinate-invariant distortion measures. 

For a scaled isometry, the condition (\ref{eq:scalediso_condition}) is equivalent to 
$\lambda_i(z) = c$ for some positive scalar $c$ and for all $z\in\mathbb{R}^{m}$. 
A coordinate-invariant {\it relaxed distortion measure} has been introduced in~\cite{lee2022regularized}:
\begin{equation}
\label{eq:rdm}
   \int_{\mathbb{R}^{m}} \sum_{j=1}^{m} (1 - \frac{\lambda_j(z)}{\int_{\mathbb{R}^{m}} \sum_{i=1}^{m} \lambda_i(z)/m \ d \nu})^2 \ d \nu,
\end{equation}
which measures how far the mapping $f$ from being a scaled isometry.  
The relaxed distortion measure is zero if and only if $f$ is any scaled isometry. 
We refer to~\cite{lee2022regularized} for a more general family of coordinate-invariant 
relaxed distortion measures.

\section{Geometric Regularization of Autoencoders}
In this section, to maintain brevity in the paper's length, we focus on explaining algorithms with minimal examples; we refer to the original papers for more experimental results.

We use the decoder Jacobian $J_f=\frac{\partial f}{\partial z}$ and sometimes its second-order derivatives 
$\frac{\partial^2 f}{\partial z^i \partial z^j}$ in the following algorithms.
As we usually use deep neural networks for $f$, computing the entire Jacobian and the second-order derivatives of $f$ during training becomes impractical due to the substantial memory and computational costs. In practice, we develop algorithms that only require the use of the Jacobian-vector and vector-Jacobian products, which can be done much more efficiently. 
For this purpose, we often use Hutchinson's stochastic trace estimator~\cite{hutchinson1989stochastic}:
\begin{equation}
    \mathrm{Tr}(A) \approx \mathbb{E}_{v \sim {\cal N}(0, I)}[v^T A v],
\end{equation}
where ${\cal N}(0, I)$ is the standard normal distribution. 

Throughout we consider a deterministic autoencoder with an encoder function
$g_{\phi}:\mathbb{R}^{D}\to\mathbb{R}^{m}$ and decoder function
$f_{\theta}:\mathbb{R}^{m}\to\mathbb{R}^{D}$, with their composition
denoted by $F_{\theta,\phi}:=f_{\theta}\circ g_{\phi}$. We use the notation 
$\mathcal{D}:=\{x_i\in \mathbb{R}^{D}\}_{i=1}^{N}$ to denote the set of observed
data points.

\subsection{Neighborhood Reconstructing Autoencoders}
\label{sec:nrae}
In this section, we provide a high-level mathematical description of the
\textbf{Neighborhood Reconstructing Autoencoder (NRAE)}~\cite{lee2021neighborhood}, 
which exploits a priori constructed neighborhood graph to regularize the geometry 
and connectivity of the learned manifold.
In what follows we use the notation $\mathcal{N}(x)$ to denote the set of
neighborhood points of $x$, with $x$ included in $\mathcal{N}(x)$.
We begin with the following definition:

\begin{defn}
\label{def:definition1}
Let $\tilde{F}_{\theta,\phi}(\cdot;x):=\tilde{f}_{\theta}(g_{\phi}(\cdot);
g_{\phi}(x))$, where $\tilde{f}_{\theta}(\cdot;z)$ is a local quadratic (or
in some cases linear) approximation of $f_{\theta}$ at $z=(z^1,z^2,...z^m)$:
\begin{equation}
\small
\label{eq:regular22}
\tilde{f_{\theta}}(z';z) := f_{\theta}(z) + \sum_{i=1}^{m} \frac{\partial f_{
\theta}}{\partial z^i}(z) d z^i + \sum_{i,j=1}^{m} \frac{1}{2}\frac{\partial^2 f_{
\theta}}{\partial z^i \partial z^j} (z) d z^i d z^j, 
\end{equation}
where $dz=z'-z$.  $\tilde{F}_{\theta,\phi}(\mathcal{N}(x);x)$ is said to be a
{\bf neighborhood reconstruction} of $\mathcal{N}(x)$.
\end{defn}

\begin{figure}[!t]
\centering
\includegraphics[width=1\linewidth]{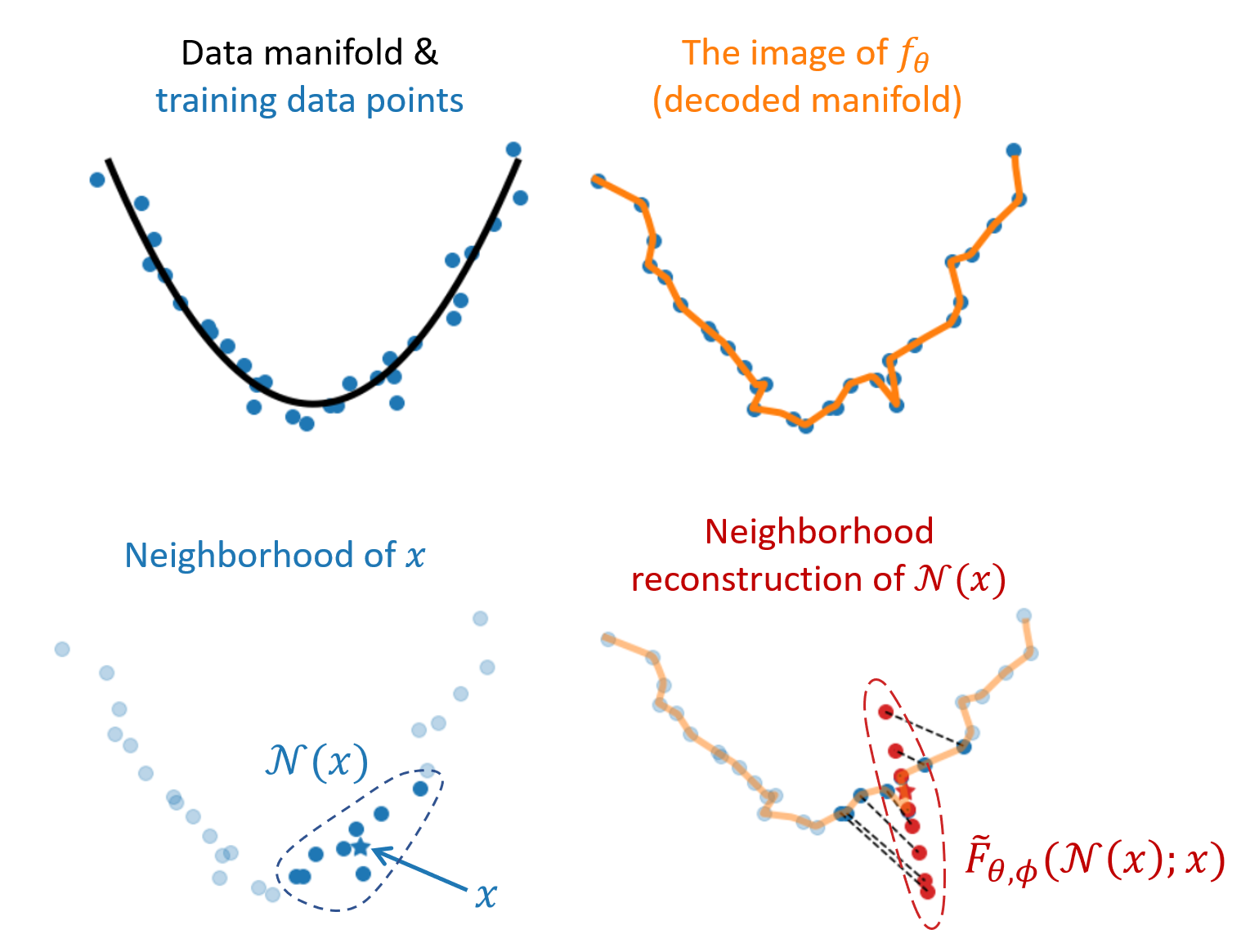}
\vspace{-10pt}
\caption{The training data points (blue), the decoded manifold (orange), the
neighborhood of $x$ denoted by $\mathcal{N}(x)$, and the neighborhood reconstruction
(red). The black dotted lines represent the correspondences between $x' \in 
\mathcal{N}(x)$ and $\tilde{F}_{\theta,\phi}(x';x)$.}
\label{fig:NRAE_main3}
% \vskip -0.5em
\end{figure}

The key idea behind Definition~\ref{def:definition1} is that we locally approximate
the decoder, and not the encoder, to extract local geometric
information on the decoded manifold, which is captured in the image of
$\tilde{F}_{\theta,\phi}(\cdot;x)$.  Figure~\ref{fig:NRAE_main3} illustrates an example where the
autoencoder reconstructs the points almost perfectly, but the neighborhood
reconstruction of $\mathcal{N}(x)$, whose elements lie in the tangent space
(here we use the linear approximation of $f_{\theta}$), is considerably
different from $\mathcal{N}(x)$.

\begin{figure}[!t]
\centering
\includegraphics[width=1\linewidth]{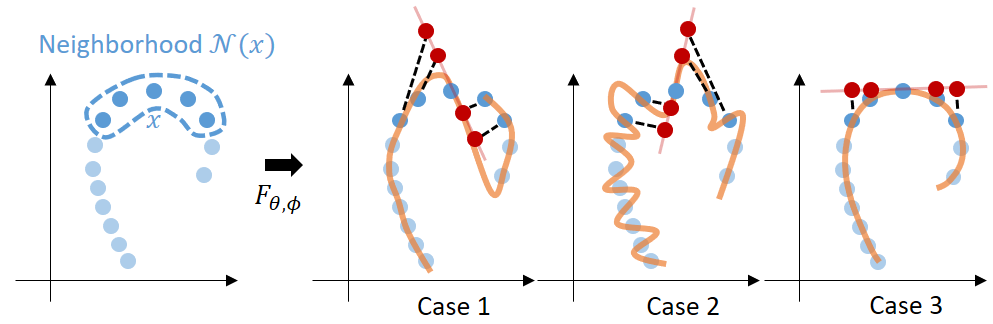}
\caption{The orange curves represent the learned manifolds, the red points 
represent the neighborhood reconstruction, and the lengths of the black-dotted
lines represent the neighborhood reconstruction loss.}
\label{fig:NRAE_main}
\end{figure}

Given that the neighborhood reconstruction of $\mathcal{N}(x)$ reflects
the local geometry of the decoded manifold, minimizing a loss function that
measures the difference between $\mathcal{N}(x)$ and its image
$\tilde{F}_{\theta,\phi}(\mathcal{N}(x);x)$ is one means of training an
autoencoder to preserve the local geometry of the original data distribution.
With that goal in mind, we formulate a {\bf neighborhood reconstruction loss}
$\mathcal{L}$ as follows:
\begin{equation}
\scriptsize
\mathcal{L}(\theta,\phi;\mathcal{D}) = \frac{1}{|\mathcal{D}|}\sum_{x\in \mathcal{D}}\frac{1}{|\mathcal{N}(x)|}\sum_{x'\in \mathcal{N}(x)} K(x',x) \cdot \|x' - \tilde{F}_{\theta,\phi}(x';x)\|^2,
\end{equation}
where $K(x',x)$ is a positive symmetric kernel function that determines
the weight for each $x'\in \mathcal{N}(x)$.  
We note that the computations in $\tilde{F}_{\theta, \phi}$ involving derivatives of $f_{\theta}$
can be done by Jacobian-vector and vector-Jacobian products.

Figure~\ref{fig:NRAE_main}
illustrates how the neighborhood reconstruction loss can differentiate among
the quality of the learned manifolds whose point reconstruction losses
are all the same (close to zero): Case 3 has the smallest neighborhood
reconstruction loss compared to Case 1 (wrong local geometry) and Case 2
(overfitting).

\begin{figure}[!t]
    \centering
    \includegraphics[width=0.9\linewidth]{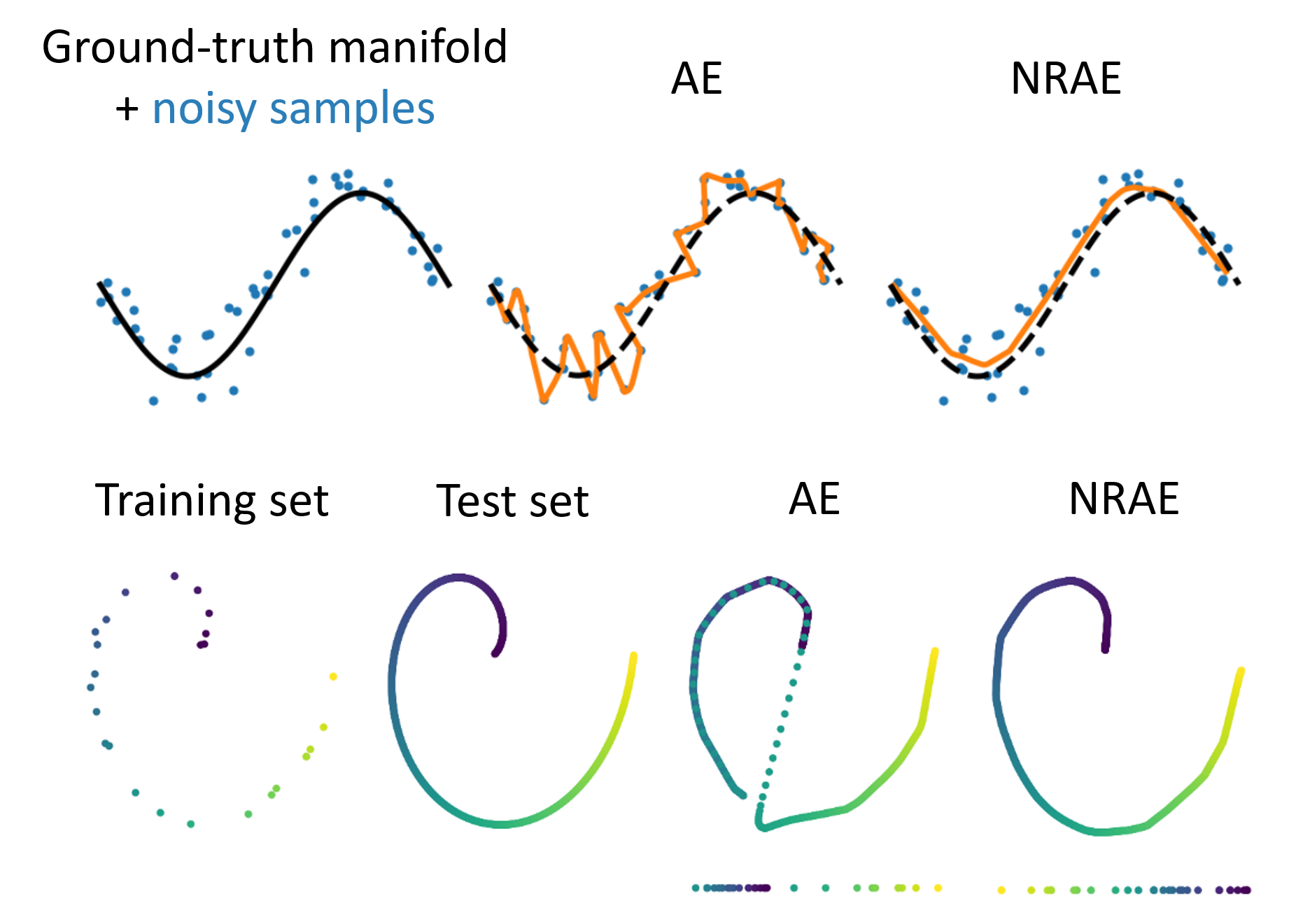}
    \caption{\textit{Upper}: Given corrupted training data points, NRAE produces more smooth manifold. \textit{Lower}: Given sparse training data points, NRAE produces a manifold with correct local connectivity (dots below the manifolds represent the one-dimensional latent representations of the training data points).}
    \vspace{-10pt}
    \label{fig:nraes}
\end{figure}
Figure~\ref{fig:nraes} shows some example results; NRAEs (with quadratic approximations) 
produce more smooth and accurate manifolds with correct connectivity compared to the vanilla AE.

\subsection{Minimum Curvature Autoencoders}
\label{sec:mcae}
In this section, we introduce the extrinsic curvature regularization method for autoencoders
~\cite{lee2023explicit}, that can resolve the wrong manifold issue to some extent. 
Recall the local extrinsic curvature measure (\ref{eq:excurvature}) is 
\begin{equation*}
    \sum_{i,j} \big((J^T_{f_{\theta}} H J_{f_{\theta}})^{-1}\big)_{ij} \mathrm{Tr}\big[ \frac{d\hat{T}}{dz^i} \frac{d\hat{T}}{dz^j} \big],
\end{equation*}
where $\hat{T} = J_{f_{\theta}} (J_{f_{\theta}}^T J_{f_{\theta}})^{-1} J_{f_{\theta}}^{T}$. 
To approximate the trace, consider a set of Gaussian samples $\{w_k \sim {\cal N}(0, I_{D\times D})\}_{k=1}^{K}$, 
and use Hutchinson’s trace estimator:
\begin{equation*}
    \sum_{i,j} \big((J^T_{f_{\theta}} H J_{f_{\theta}})^{-1}\big)_{ij} \frac{1}{K}\sum_{k=1}^{K} \big[ \frac{d (w_k^T\hat{T})}{dz^i} \frac{d(\hat{T} w_k)}{dz^j} \big].
\end{equation*}
This can be re-written by using the trace as follows:
\begin{equation*}
    \frac{1}{K} \sum_{k=1}^{K} \mathrm{Tr} \big[ (J^T_{f_{\theta}} H J_{f_{\theta}})^{-1}
    (\frac{d (\hat{T} w_k)}{dz})^T \frac{d(\hat{T} w_k)}{dz} \big],
\end{equation*}
where $\frac{d(\hat{T} w_k)}{dz}$ is a $D\times m$ matrix. 
We can again use a set of Gaussian samples $\{v_l \sim {\cal N}(0, I_{m\times m })\}_{l=1}^{L}$, and use the trace estimator:
\begin{equation*}
    \frac{1}{K} \sum_{k=1}^{K} \frac{1}{L} \sum_{l=1}^{L} v_l^T (J^T_{f_{\theta}} H J_{f_{\theta}})^{-1}
    (\frac{d (\hat{T} w_k)}{dz})^T \frac{d(\hat{T} w_k)}{dz} v_l.
\end{equation*}
In practice, it is sufficient to use one sample in each trace estimation. 
Because of the matrix inverse computation, we need to compute the full Jacobian, 
but other computations can be done by Jacobian-vector or vector-Jacobian products. 

\begin{figure}[!t]
    \centering
    \includegraphics[width=0.9\linewidth]{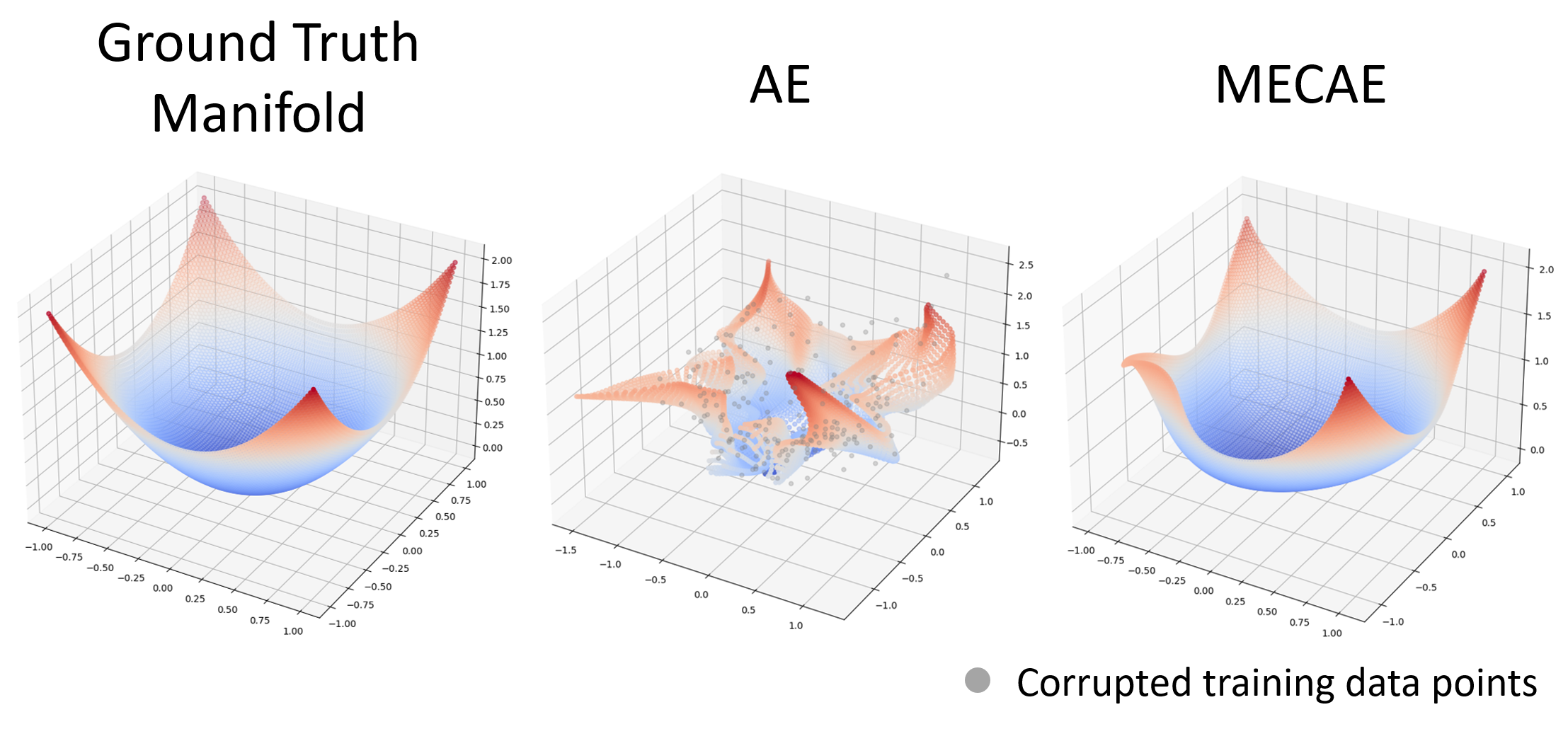}
    \caption{Given corrupted training data points, MECAE -- where $H(x)=I_{3\times 3}$ -- produces a more smooth and accurate manifold than the AE.}
    \label{fig:mcaes}
\end{figure}

This is a local (approximate) extrinsic curvature measure, and we want to minimize the curvature of the manifold globally. Denoting the local extrinsic curvature measure by $\text{LEC}(z)$, we compute the global curvature measure with respect to some positive measure $\nu$ as follows: $\int_{\mathbb{R}^{m}} \text{LEC}(z) \ d\nu(z).$
In practice, we construct a probability measure $\nu$ by using the encoder function. 
Let $p_{\phi}(z)$ be a probability density function defined as the push-forward of the 
data distribution, then it is approximated by an empirical distribution 
$\frac{1}{|{\cal D}|} \sum_{i} \delta(z-g_{\phi}(x_i))$. Using this as $\nu$, 
the global curvature measure can be approximately computed by $\frac{1}{|{\cal D}|} \sum_{x \in {\cal D}} \text{LEC}(g_{\phi}(x)).$
This is added to the original reconstruction loss function with a suitable weighting parameter $\alpha$:
\begin{equation}
    \frac{1}{|{\cal D}|} \sum_{x \in {\cal D}} \big(\|F_{\theta, \phi}(x) - x\|^2 + \alpha \ \text{LEC}(g_{\phi}(x)) \big),
\end{equation}
which is the final objective function for the \textbf{Minimum Extrinsic Curvature Autoencoder (MECAE)}; the intrinsic curvature minimization method is also included in~\cite{lee2023explicit}. Figure~\ref{fig:mcaes} 
shows an example result; MECAE produces a smoother and more accurate manifold.

\subsection{Isometrically Regularized Autoencoders}
\label{sec:irae}
In this section, we explain the isometric regularization method for autoencoders, 
which attempts to minimize the geometric distortion in the latent space 
by regularizing the decoder to be a scaled isometry~\cite{lee2022regularized}.  
Recall that the relaxed distortion measure (\ref{eq:rdm}) is 
\begin{equation*}
   \int_{\mathbb{R}^{m}} \sum_{j=1}^{m} (1 - \frac{\lambda_j(z)}{\int_{\mathbb{R}^{m}} \sum_{i=1}^{m} \lambda_i(z)/m \ d \nu})^2 \ d \nu,
\end{equation*}
where $\{\lambda_i(z)\}=\text{eigenvalues}(J_{f_\theta}^T(z)H(f(z)) J_{f_\theta}(z)$). 
Replacing $\int_{\mathbb{R}^{m}} \cdot \ d\nu$ by $\mathbb{E}_{z \sim P}$ with respect to some 
probability measure $P$, we can re-write the expression as follows:
\begin{equation*}
    m^2\frac{\mathbb{E}_{z \sim P}[\mathrm{Tr}(
        (J_{f_{\theta}}^T H J_{f_{\theta}})^2) 
    ]}{\mathbb{E}_{z\sim P}[\mathrm{Tr}(J_{f_{\theta}}^T H J_{f_{\theta}})]^2} - m.
\end{equation*}

Then, we can use Hutchinson’s trace estimator with Gaussian samples $v,w \sim {\cal N}(0, I_{m \times m})$. 
Ignoring the constant multiple $m^2$ and the shift $-m$, the approximate relaxed distortion measure is 
\begin{equation*}
   \frac{\mathbb{E}_{z \sim P}\big[\mathbb{E}_{v}[v^T
        (J_{f_{\theta}}^T H J_{f_{\theta}})^2 v]\big]}{
        \mathbb{E}_{z\sim P}\big[\mathbb{E}_{w} [w^T J_{f_{\theta}}^T H J_{f_{\theta}} w]\big]^2},
\end{equation*}
which can be computed by using the Jacobian-vector and vector-Jacobian products efficiently. 

\begin{figure}[!t]
    \centering
    \includegraphics[width=\linewidth]{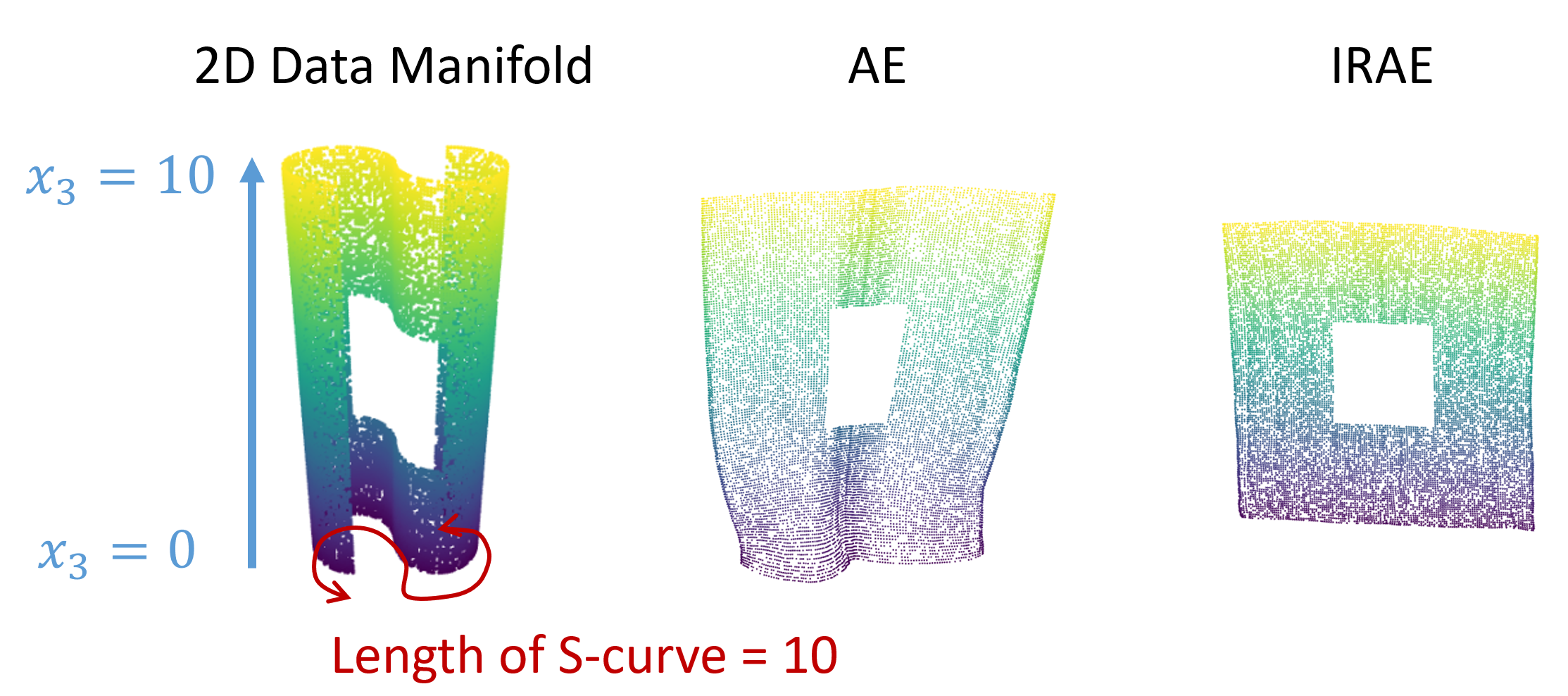}
    \caption{Given a two-dimensional manifold (square-shape manifold with a square hole) in $\mathbb{R}^{3}$, the IRAE -- where $H(x)=I_{3\times 3}$ -- produces a geometry-preserving latent space, while AE produces distorted latent space.}
    \vspace{-10pt}
    \label{fig:iraes}
\end{figure}

We construct a probability measure $P$ by using the encoder function; 
$p_{\phi}(z)$ is approximated by an empirical distribution 
$\frac{1}{|{\cal D}|} \sum_{i} \delta(z-g_{\phi}(x_i))$. 
Then, the relaxed distortion measure can be approximately computed.
This is added to the original reconstruction loss function with a suitable weighting parameter $\alpha$:
\begin{equation}
\small
    \frac{1}{|{\cal D}|} \sum_{x \in {\cal D}} \big(\|F_{\theta, \phi}(x) - x\|^2 + \alpha \ \frac{\mathbb{E}_{v}[v^T
        (J_{f_{\theta}}^T H J_{f_{\theta}})\big|_{g_{\phi}(x)}^2 v]}{
        \mathbb{E}_{w}[w^T (J_{f_{\theta}}^T H J_{f_{\theta}})\big|_{g_{\phi}(x)} w]^2} \big),
\end{equation}
where $v,w$ are samples from ${\cal N}(0, I_{m\times m})$. In practice, we use one sample for 
$v$ and $w$, and let $v=w$ for more efficient computation. This is the final loss function for the \textbf{Isometrically Regularized Autoencoder (IRAE)}~\cite{lee2022regularized}. 
Figure~\ref{fig:iraes} shows an example result, where IRAE produces a geometry-preserving latent space while AE produces 
distorted latent space.

As discussed in~\cite{lee2023explicit}, according to Gauss's Theorema Egregium (i.e., Gauss's Remarkable Theorem), the Gaussian curvature of a surface is invariant under local isometry. In other words, if two surfaces or manifolds are mapped to each other without distortion, then their Gaussian curvatures or intrinsic curvatures should be preserved. 
Therefore, if $f$ is a scaled isometry, then since the intrinsic curvature of the Euclidean latent space is everywhere 0, the resulting manifold's intrinsic curvature must be 0 everywhere as well.
As a result, in IRAE, intrinsic curvatures are implicitly minimized as a byproduct of distortion minimization. 

In all experiments in~\cite{lee2022regularized}, the ambient space metric $H$ is assumed to be the identity metric $I_{D\times D}$. 
There is one specific work that does not use the identity metric in isometric regularization.
\citet{lee2022statistical} have considered the case where the ambient data space is a point cloud manifold, 
where each point is a point cloud, and have developed the \textbf{Info-Riemannian metric} for the point cloud manifold based on the theory of information geometry and statistical manifold~\cite{amari2016information, amari2000methods}.   

\section{Conclusions}
The geometric understanding, or the manifold learning viewpoint, of the autoencoder framework 
gives us good insights into what autoencoders actually learn and what problems exist in vanilla autoencoders. In the beginning, we have shown that there are two fundamental issues in the 
training of autoencoders via reconstruction loss alone: (i) wrong manifold issue and (ii) 
distorted latent space issue. 
Then we have introduced three recent works, NRAE~\cite{lee2021neighborhood}, MCAE~\cite{lee2023explicit}, and IRAE~\cite{lee2022regularized}, that address these issues.  
Commonly, these methods focus on the geometric characteristics of the decoder $f$ 
by investigating its Jacobian and higher-order derivatives. 
Considering the memory and computation costs of the derivatives of deep neural networks, 
practical approximation formulas have been given, relying on the Jacobian-vector and vector-Jacobian 
products.

In most existing studies, along with the previously mentioned studies, 
the ambient space metric $H(x)$ was assumed to be identity $I$. 
The choice of the metric will significantly affect the resulting manifold 
and representations, and yet, relatively little attention has been given 
to this problem so far. We believe investigating this aspect further 
in future research holds promise and is worth exploring.

\bibliography{example_paper}
\bibliographystyle{icml2022}

% \newpage
% \appendix
% \onecolumn
% % \section{You \emph{can} have an appendix here.}

% You can have as much text here as you want. The main body must be at most $8$ pages long.
% For the final version, one more page can be added.
% If you want, you can use an appendix like this one, even using the one-column format.
%%%%%%%%%%%%%%%%%%%%%%%%%%%%%%%%%%%%%%%%%%%%%%%%%%%%%%%%%%%%%%%%%%%%%%%%%%%%%%%
%%%%%%%%%%%%%%%%%%%%%%%%%%%%%%%%%%%%%%%%%%%%%%%%%%%%%%%%%%%%%%%%%%%%%%%%%%%%%%%

\end{document}